\documentclass[journal,10pt]{IEEEtran}

\usepackage{cite}
\usepackage{amsmath,amssymb,amsfonts}
\usepackage{graphicx}
\usepackage{textcomp}
\usepackage{hyperref}
\usepackage{xcolor}
\usepackage{soul,color}
\usepackage{url}
\usepackage{balance}
\usepackage[ruled,vlined,linesnumbered]{algorithm2e}
\def\BibTeX{{\rm B\kern-.05em{\sc i\kern-.025em b}\kern-.08em
    T\kern-.1667em\lower.7ex\hbox{E}\kern-.125emX}}
\begin{document}

\title{Genetically Optimized Prediction of Remaining Useful Life}

\author{\IEEEauthorblockN{Shaashwat Agrawal\IEEEauthorrefmark{1},
Sagnik Sarkar\IEEEauthorrefmark{1},Gautam Srivastava*\IEEEauthorrefmark{2}, \\Praveen Kumar Reddy Maddikunta,\IEEEauthorrefmark{3} Thippa Reddy Gadekallu\IEEEauthorrefmark{3}}\\
\IEEEauthorblockA{
\IEEEauthorrefmark{1}School of Computer Science and Engineering, Vellore Institute of Technology, Vellore, India\\
E-mail:shaas2000@gmail.com,sagnik106@gmail.com\\
\IEEEauthorrefmark{2}  Department of Mathematics and Computer Science, Brandon University, Manitoba, Canada.\\\textcolor{blue}{E-mail: srivastavag@brandonu.ca}\\
\IEEEauthorrefmark{3} School of Information Technology and Engineering, Vellore Institute of Technology, Vellore, India \\E-mail: praveenkumarreddy@vit.ac.in, thippareddy.g@vit.ac.in
}}
		

 
 
 
 
\maketitle

\begin{abstract}
The application of remaining useful life (RUL) prediction has taken great importance in terms of energy optimization, cost-effectiveness, and risk mitigation. The existing RUL prediction algorithms mostly constitute deep learning frameworks. In this paper, we implement LSTM and GRU models and compare the obtained results with a proposed genetically trained neural network. The current models solely depend on Adam and SGD for optimization and learning. Although the models have worked well with these optimizers, even little uncertainties in prognostics prediction can result in huge losses. We hope to improve the consistency of the predictions by adding another layer of optimization using Genetic Algorithms. The hyper-parameters - learning rate and batch size are optimized beyond manual capacity. These models and the proposed architecture are tested on the NASA Turbofan Jet Engine dataset. The optimized architecture can predict the given hyper-parameters autonomously and provide superior results.
\end{abstract}

\begin{IEEEkeywords}
LSTM, GRU, genetically trained neural network, prognostic, hyper-parameters,learning rate, batch size, Remaining Useful Life.
\end{IEEEkeywords}

\section{Introduction}
\label{intro}
Remaining Useful Life is an important characteristic of any machinery or battery. Engines, lithium-ion batteries, water pumps are some such machinery that requires constant maintenance as their efficiency decreases with time. Prediction of their useful life helps industries in easy replacement, cost-effectiveness, and production efficiency by changing from systematic to condition-based maintenance \cite{maddikunta2020predictive}. Various datasets like NASA turbofan jet engine, lithium-Ion battery have been used in the research of this field \cite{zhang2018long,ghorbani2020estimating}. 

The NASA turbofan jet engine dataset is a sequential dataset that inspects a variety of aspects required for prediction. It measures 3 operational parameters and 21 sensor values. Various model-based and data-driven approaches have been applied to study and understand this data. Most research comes under statistical-based methods of Dynamic Bayesian Networks \cite{medjaher2012remaining} and machine learning algorithms. Deep Learning models of Long-Short Term Memory(LSTM) and Gated Recurrent Unit(GRU) have been used to study this dataset extensively. Given its time-series data, these models have shown significant results in this field. Various hybrid layers have also been included to compensate for the complexity of the basic recurrent models and build on them. These mainly include 1-D, 2-D Convolution layers, and  Bi-directional, Multi-Directional LSTM layers\cite{li2018remaining,li2019directed,9079864}. 

In any RUL application, precision takes huge priority to avoid accidents and huge losses. Even if composite algorithms and models can provide accurate results, consistency is important. In any particular unit time cycle, there can be various possible outcomes depending on even a slight change in operational settings and other factors. Research on optimization has increased to cover this necessity. All Deep Learning architectures utilize a certain optimizer for their learning but these alone might not be enough. In general, Stochastic Gradient Descent(SGD) and Adam are used due to their efficiency and consistency. 

Other optimization methodologies are used on top of the default ones to get better results. Adaboost, Recursive Levenberg-Marquardt (RLM) are some algorithms used. Following them, evolutionary algorithms have also become known for their optimization success. Nature-inspired algorithms are being used extensively in various ways to minimize the energy and optimize the training of neural networks \cite{maddikunta2020green}. Their flexibility in terms of application opens up many possibilities in terms of optimization. They can be used in deep learning to optimize hyper-parameters, activation functions, model architectures, and so on. 

In this paper, the NASA dataset is trained with a tuned 2-layer LSTM and GRU models. The results are compared to existing outputs and inference shown. After consulting a lot of literature, we introduce a semi-novel optimization algorithm using Genetic Algorithms. Hyper-parameters of the model - learning rate and batch size are self tuned by a Genetic Algorithm for every generation. $\Delta$ validation loss of every individual in a generation is taken and evaluated. Top individuals pass on their genes to the next generation. This methodology of optimization works hand in hand with Adam to prevent over-fitting and under-fitting of data. In Section \ref{relatedwork}, the domains under study are defined. Section \ref{arch} introduces the existing LSTM and GRU model architectures in use and the proposed training methodology using Genetic Algorithms. After this definition, the results of the implemented architectures and their comparisons are shown in Section \ref{results}. Section \ref{Conclusion} deals with conclusion and future work.

\section{Literature Survey}
\label{relatedwork}

\subsection{Remaining Useful Life Prediction}
Degradation and rusting occur in every component and element of our environment. As research progresses, we tend to move from the discipline of diagnosis to one of prognosis. Similarly, instead of systematic and continuous maintenance, the current research focuses on predictive and condition-based maintenance. The remaining Useful Life of any machinery can be predicted by a recorded history of operational conditions, various sensor values, and other evaluation metrics. Instead of focusing on physical models, \cite{ghorbani2020estimating} that prevent failure, data-driven machine learning and other algorithmic models are preferred \cite{medjaher2012remaining,si2011remaining}. This is because the ease and efficiency of data-driven models have attracted more attention than other methodologies.

\subsection{Machine Learning and Recurrent Neural Networks}
Recurrent Neural Networks get their success from the sequential nature of data. Simple RNNs, LSTMs \cite{deng2020remaining,wu2018remaining}, Bi-directional LSTMs, and GRUs are the most utilized recurrent deep learning networks. The nature of the remaining useful life data makes them the perfect statistical tool. They can analyze minute trends in sensor values and still remember long term details. LSTMs occupy the center stage of these recurrent networks because of their structure. Convolutional layers and CNN-LSTM layers have also been stacked on existing LSTM layers for RUL prediction to increase accuracy. Depending on data pre-processing and dimension of convolution, the data attributes can be treated as features of an image and extracted similarly \cite{gadekallu2020deep}. On close observation of the data, it can be inferred that the value ranges provided by each sensor do not amount to much, especially considering the mode. Feature extraction in these cases has to be tuned carefully or it could result in the exploding gradient problem. To overcome similar issues, complex training architectures like auto-encoders have also been proposed \cite{P1}. 
 
\subsection{Evolutionary Algorithms and Optimization}

Adam and SGD optimizers are commonly used for training any deep learning architecture. They enable the model to learn through a loss function using back-propagation. At times these optimizations alone are not enough. Adaboost, Recursive Levenberg-Marquardt, and evolutionary algorithms can be used together with the default ones to better the training process. The flexibility and compatibility of Genetic Algorithms with deep learning frameworks have incited huge amounts of research. Traditional Genetic Algorithm, Artificial Bee Colony Algorithm \cite{kumar2014novel} and NEAT Algorithm \cite{elsaid2019ant} are some extensively used optimizers over neural networks. Genetically optimized LSTM models are also used for similar tasks of stock prediction \cite{chung2018genetic} and water temperature prediction \cite{stajkowski2020genetic}.

\section{Preliminaries and Proposed Architecture}\label{arch}

\begin{figure}[h!]
	\centering
	\includegraphics[width=1\linewidth]{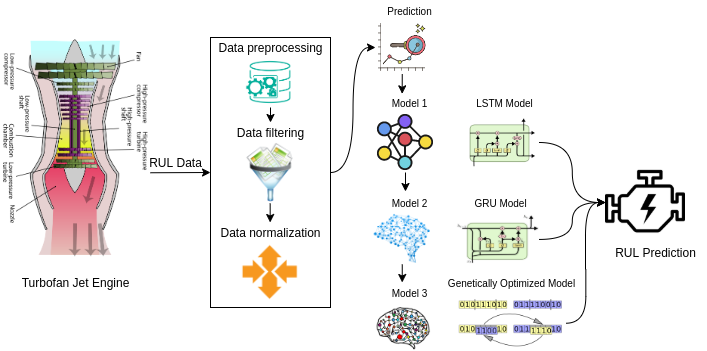}
	\caption{Structure Diagram of Implementation Models.}
	\label{fig:structure}
\end{figure}

In this section, we discuss three different architectures: LSTM, GRU, and Genetically Optimized-LSTM respectively. The subsections \ref{lstmarchi} and \ref{gruarchi} discuss the detailed architecture of the generic LSTM and GRU models and their training parameters. Subsection \ref{proparchi} discusses in detail the methodology followed in the proposed training model and the specific parameters of fitness criteria. 
The Turbofan jet engine \cite{dataset} dataset is processed, filtered, and normalized. The models are trained on input data of shape, (timeSteps, features) and are later validated. A set of 10 well-defined individuals are trained in each generation to tune the learning rate and batch size. Figure \ref{fig:structure} shows the methodology adopted in this paper.

\subsection{LSTM Architecture}\label{lstmarchi}

\begin{figure}[h!]
	\centering
	\includegraphics[width=.85\linewidth]{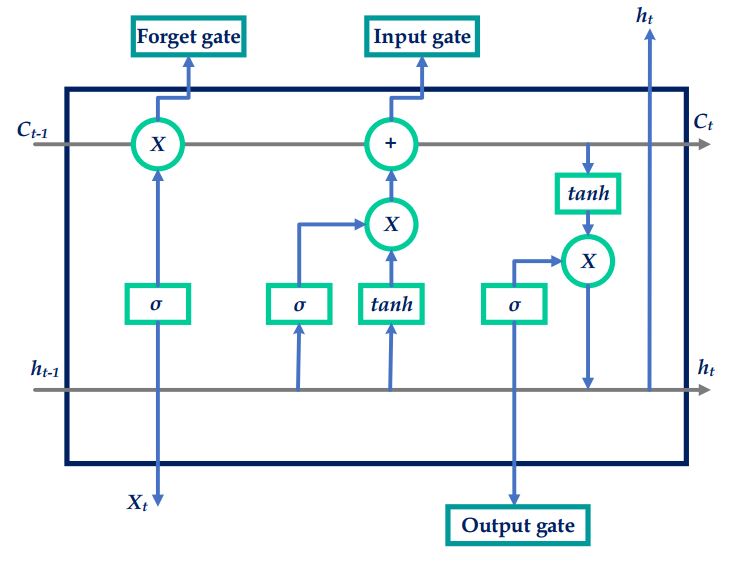}
	\caption{Long Short Term Memory Cell.}
	\label{fig:lstm}
\end{figure}

The implementation is a 3-layer model with two LSTM layers and a single dense layer. Each LSTM layer consists of 64 neurons which are followed by a 1-neuron output layer. The first LSTM layer returns sequences to the successive LSTM layer. The output layer passes through a ReLU activation function for eliminating negative prediction of life and leave the positive output unrectified. The cost functions Mean Squared Error(MSE) and Mean Absolute Error(MSE) are used for evaluating the model for both training and validation. We have chosen this cost function due it its performance on regression models.
In Figure \ref{fig:lstm}, each LSTM cell consists of a forget gate, input gate, and output gate which helps it have long-term memory by default \cite{parimalaspatiotemporal}.

\subsection{GRU Architecture}\label{gruarchi}

The implementation of the GRU model is very similar to the LSTM model. The GRU model has 3 layers - 2 GRU layers each consisting of 64 parallel GRU cells and followed by a single 1 neuron wide output layer. The first GRU layers return sequences to the second GRU layer which unlike the previous layer does not return a sequence. The output of this second GRU layer then is connected to the output layer which uses a ReLU activation for rectification of the output and preventing negative values as the predicted life. The same cost functions Mean Squared Error(MSE) and Mean Absolute Error(MAE) are used to evaluate the model on train and validation data due to its high performance on regression models.

\subsection{Proposed Architecture}
\label{proparchi}

\begin{figure}[h!]
	\centering
	\includegraphics[width=\linewidth]{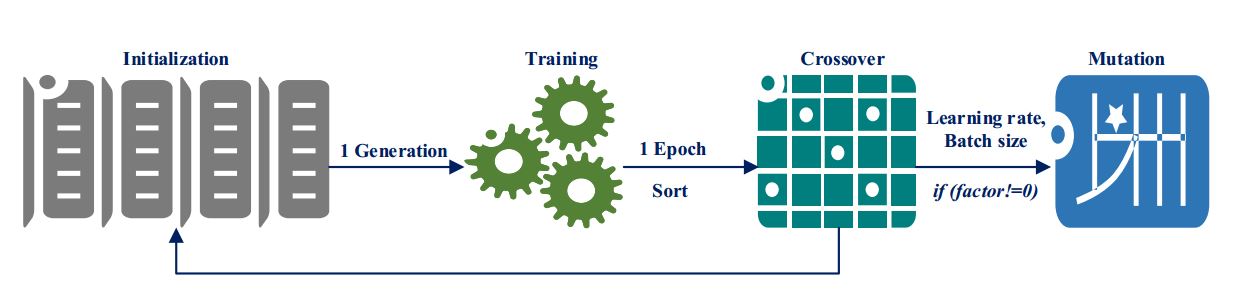}
	\caption{Proposed Architecture Model for Optimization Process.}
	\label{fig:architecture}
\end{figure}

We propose a genetic optimization methodology to tune the learning rates and batch sizes of the LSTM and GRU architectures. Fine-tuning any model requires a great deal of time and yet it is almost impossible to get the perfect learning rate required for any model. To overcome such difficulties we implement the Genetic Algorithm for each epoch of our model to tune these hyper-parameters. The methodology can be divided into some major subsections: initialization, training, cross-over, mutation. Figure \ref{fig:architecture} shows complete flow of the optimization process.

A generation is \textbf{initialized} at the beginning of training. A random set of commonly used learning rate and batch size values is parameterized to each model. For generation 1, a new set of individuals are formed. Later generations consist of individuals, and children formed by the mating of parents present in the previous one. Each generation has genes of its direct predecessor and hence performs better than them.

Every individual in a generation is \textbf{trained} for a epoch. An individual is an LSTM/GRU model with randomly initialized weights. After training their losses are stored and evaluated. The currently available batch sizes are allotted to each model in order of succession. Every model shows some decrease in their loss after training. This loss ($loss_{current}$) is stored and subtracted from the previous loss ($loss_{prev}$) to evaluate the learning that took place in that epoch. Individuals with poor $\Delta$ loss Eqn.\ref{eq:delloss} indicate the presence of improper hyper-parameters.

\begin{equation}
\Delta loss=loss_{current} - loss_{prev}\label{eq:delloss}
\end{equation}

The individuals are sorted concerning $\Delta$ loss. Top individuals are directly promoted to the next generation. The rest of the new generation is formed by \textbf{cross-over} of individuals from the previous generation. Considering any two-parent individuals at random, one out of four combinations of learning rate and batch size are selected and assigned to a cloned model as a new child. The core prosperity of any Genetic Algorithm comes from cross-over. It introduces randomness and yet completeness in the new generation. After every cross-over, a possibility of mutation is explored. A random factor Eqn.\ref{eq:factor} is chosen from (-1, 0, 1). A non-zero value will result in a 10\% mutation of the learning rate Eqn.\ref {eq:lr}. 

\begin{equation}
factor = (-1, 0, 1)\label{eq:factor}
\end{equation}
\begin{equation}
learning Rate(lr)=lr + factor\times\frac{lr}{10}\label{eq:lr}
\end{equation}

Ideally, 66.7\% of each generation undergoes \textbf{mutation}. This results in variable learning rates which would not have been possible manually. Mutation can be performed in various ways depending on application and dataset. 

\section{Results and Discussion}\label{results}

The advantage of having LSTM and GRU models for useful life prediction is the sequential nature of the data. They can identify the minute changes in sensor values and identify the trend in the data. Mean squared Error and Mean Absolute Error with cross-validation are used to measure the performance of these models. All models have been implemented and tested with TensorFlow \cite{tensor}. The tests are carried out on the NASA Turbofan Jet Engine data set containing 3 operational setting values and 21 sensor values. It is divided into cases of 250 engines each for train and test.

\subsection{Data Preprocessing}

\begin{figure*}[h!]
	\centering
	\includegraphics[width=.95\linewidth]{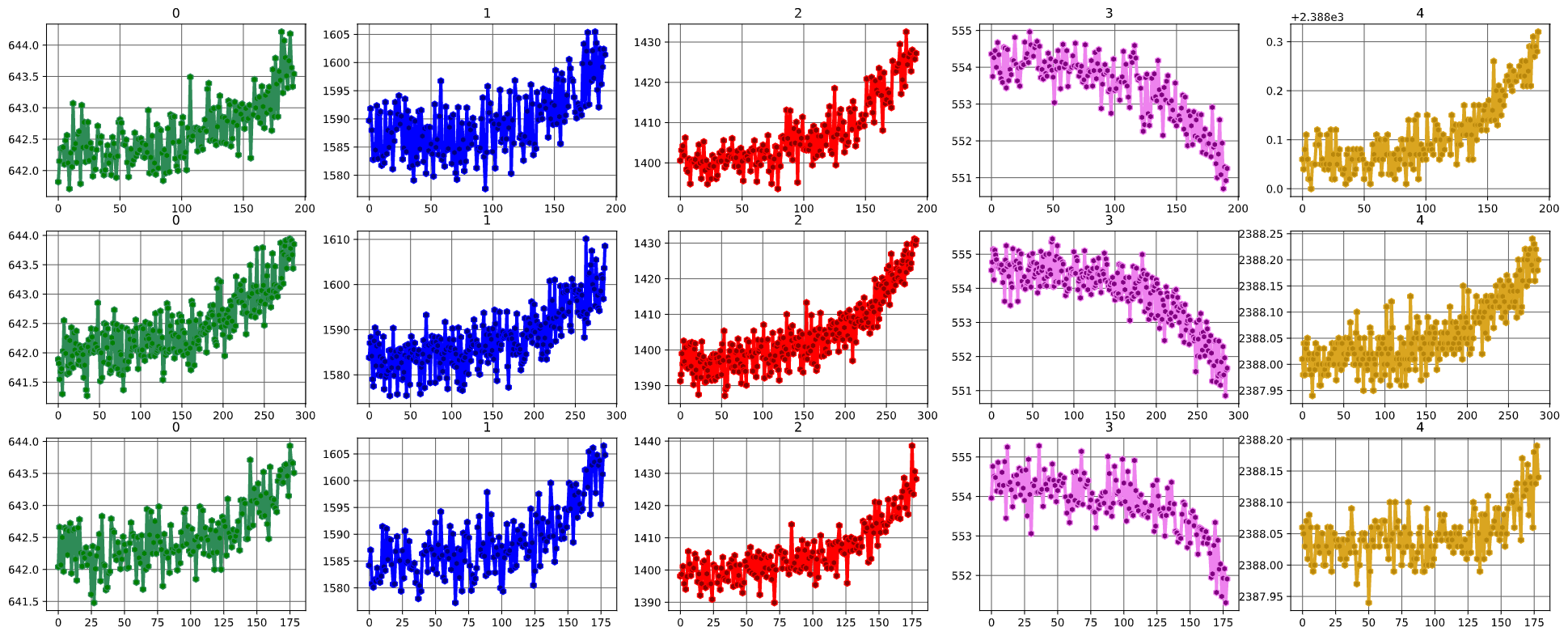}
	\caption{Plot of the time series of a few sensors belonging to an engine}
	\label{fig:time}
\end{figure*}

\begin{figure}[h!]
	\centering
	\includegraphics[width=\linewidth]{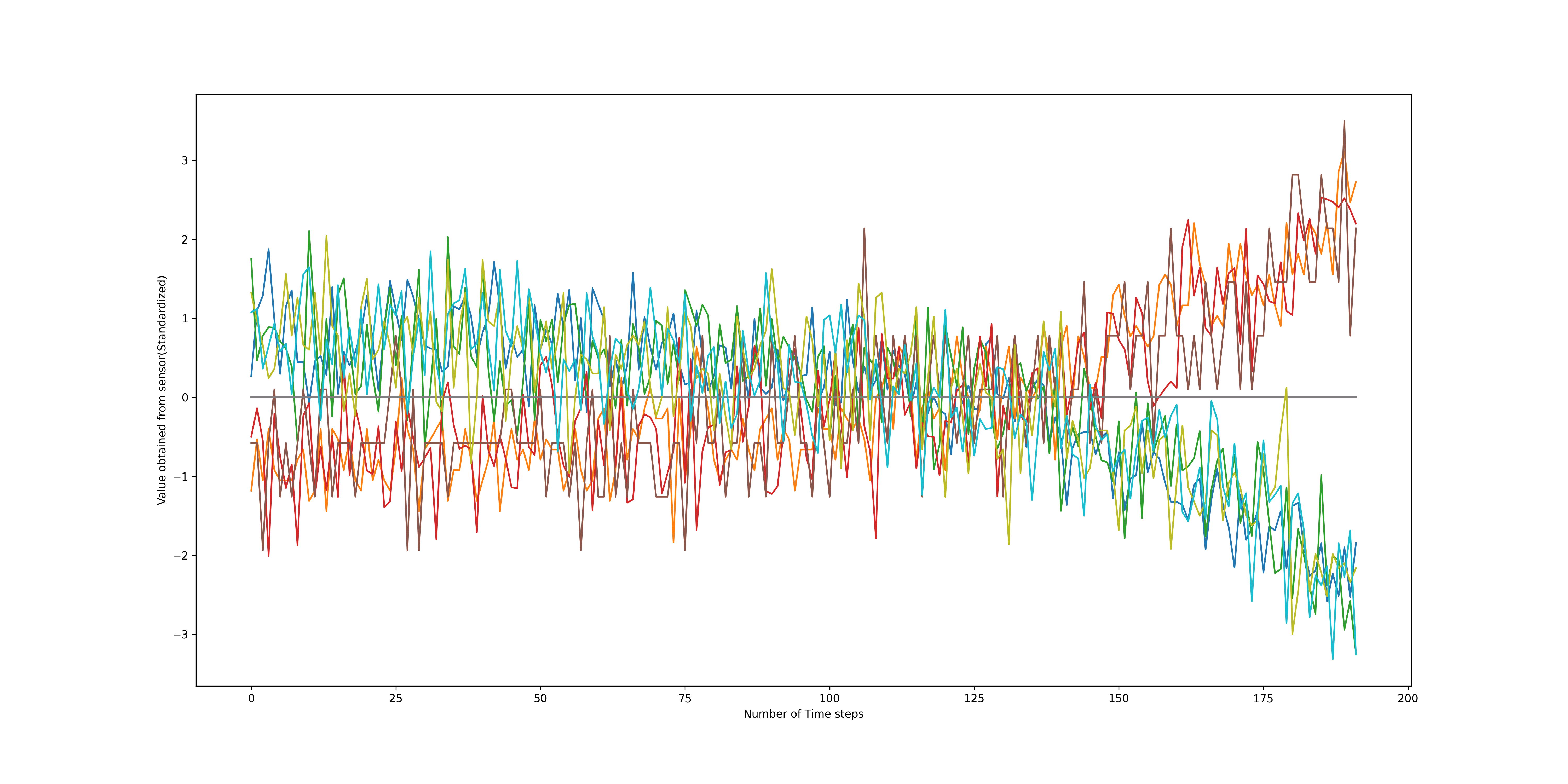}
	\caption{Standardized sensor values.}
	\label{fig:dataset}
\end{figure}

The NASA dataset helps in temporal analysis of the remaining useful life prediction. It contains a total of 26 attributes categorized into Engine number, Cycle number, 3 operational settings, and 21 sensor values. Figure \ref{fig:time} shows the time series of a few sensors for three engines. Each column in Figure \ref{fig:time} represents an engine and the rows represent particular sensors. The first 2 parameters as well as operational settings do not contribute to the predictions. Likewise, some sensor values also show constant values for changing cycles and hence hold no significant value. Each value in the dataset is normalized using Eqn.\ref{eq:norm} so that it ranges between [0,1] to avoid high variation in the data. Figure \ref{fig:dataset} shows the standardized data points for the sensor values for a single-engine.

\begin{equation}
Normalize(x_i)=\frac{x_i - min(x)}{max(x) - min(x)}\label{eq:norm}
\end{equation}

All the values of the sensors and operating modes are taken and grouped in terms of engines. The total number of time cycles is calculated by the number of entries for each engine in the dataset. 

Unlike the usual regression problems, data needs to be grouped sequentially and broken into a constant amount of time steps as mentioned to perform trend analysis. This data is then ready to be input into the model. We have chosen a time series of a total of 20 historical time steps and a prediction of the remaining number of time steps. We have thus obtained 146179 sequences of shape (20, 24) each. The expected output value is the normalized remaining life of the jet engine at any given time cycle of the engine.

\subsection{Training of LSTM model}

\begin{figure}[htbp]
	\centering
	\includegraphics[width=0.7\linewidth]{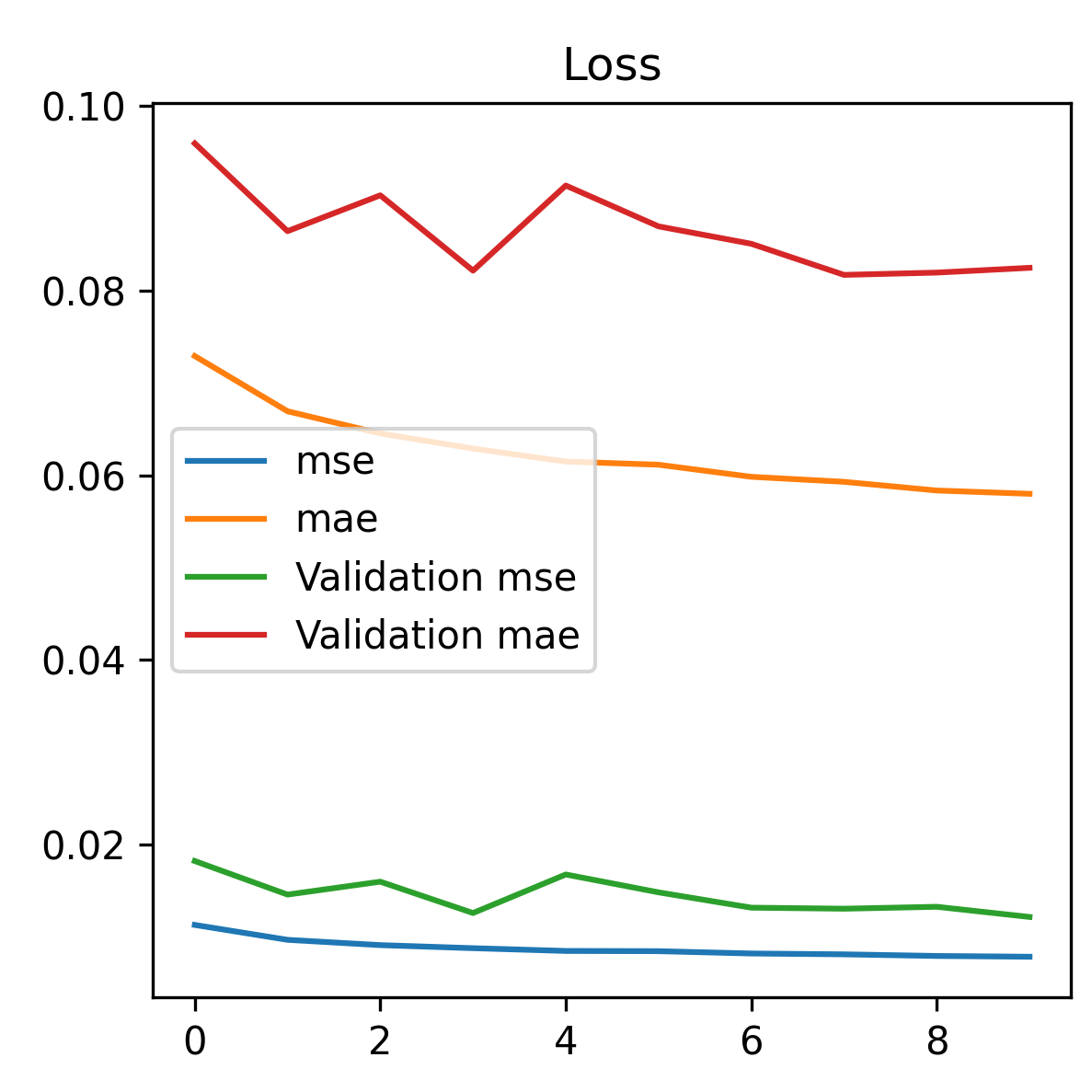}
	\caption{Performance metrics of the LSTM model.}
	\label{fig:lstmper}
\end{figure}

LSTMs (Long Short Term Memory) models are modified RNNs that can store long term memory of data. They can provide consistency of results and understand the flow of data better. We have built a 3-layer model with 2 LSTM layers and 1 dense layer. Each LSTM layer in the model contains 64 LSTM cells which mediate the flow of information. This LSTM network is trained on the Adam optimizer with a mean absolute error as the cost function for 10 epochs. We have obtained a falling curve for the training loss metric as shown in Figure \ref{fig:lstmper}. The model performance metrics are tabulated in Table \ref{table:lstm}.
\begin{table}[htbp]
\begin{center}
\centering
\begin{tabular}{|c|c|c|c|c|} 
\hline
    \textbf{Epochs} & \textbf{MSE} & \textbf{MAE} & \textbf{val. MSE} & \textbf{val. MAE}\\
    \hline
    1 & 0.0112 & 0.0729 & 0.0182 & 0.0960\\ 
    2 & 0.0096 & 0.0669 & 0.0145 & 0.0865\\ 
    3 & 0.0091 & 0.0645 & 0.0159 & 0.0903\\ 
    4 & 0.0087 & 0.0629 & 0.0125 & 0.0822\\ 
    5 & 0.0084 & 0.0615 & 0.0167 & 0.0914\\ 
    6 & 0.0084 & 0.0611 & 0.0148 & 0.0870\\
    7 & 0.0081 & 0.0598 & 0.0131 & 0.0851\\ 
    8 & 0.0081 & 0.0593 & 0.0130 & 0.0817\\ 
    9 & 0.0079 & 0.0583 & 0.0132 & 0.0820\\
    10 & 0.0078 & 0.0580 & 0.0121 & 0.0825\\
    \hline
\end{tabular}
\end{center}
\caption{The cost function(MSE - Mean Squared Error and MAE - Mean Absolute Error) for both training and validation for the LSTM model.}
\label{table:lstm}
\end{table}

\subsection{Training of GRU model}
\begin{figure}[htbp]
	\centering
	\includegraphics[width=0.7\linewidth]{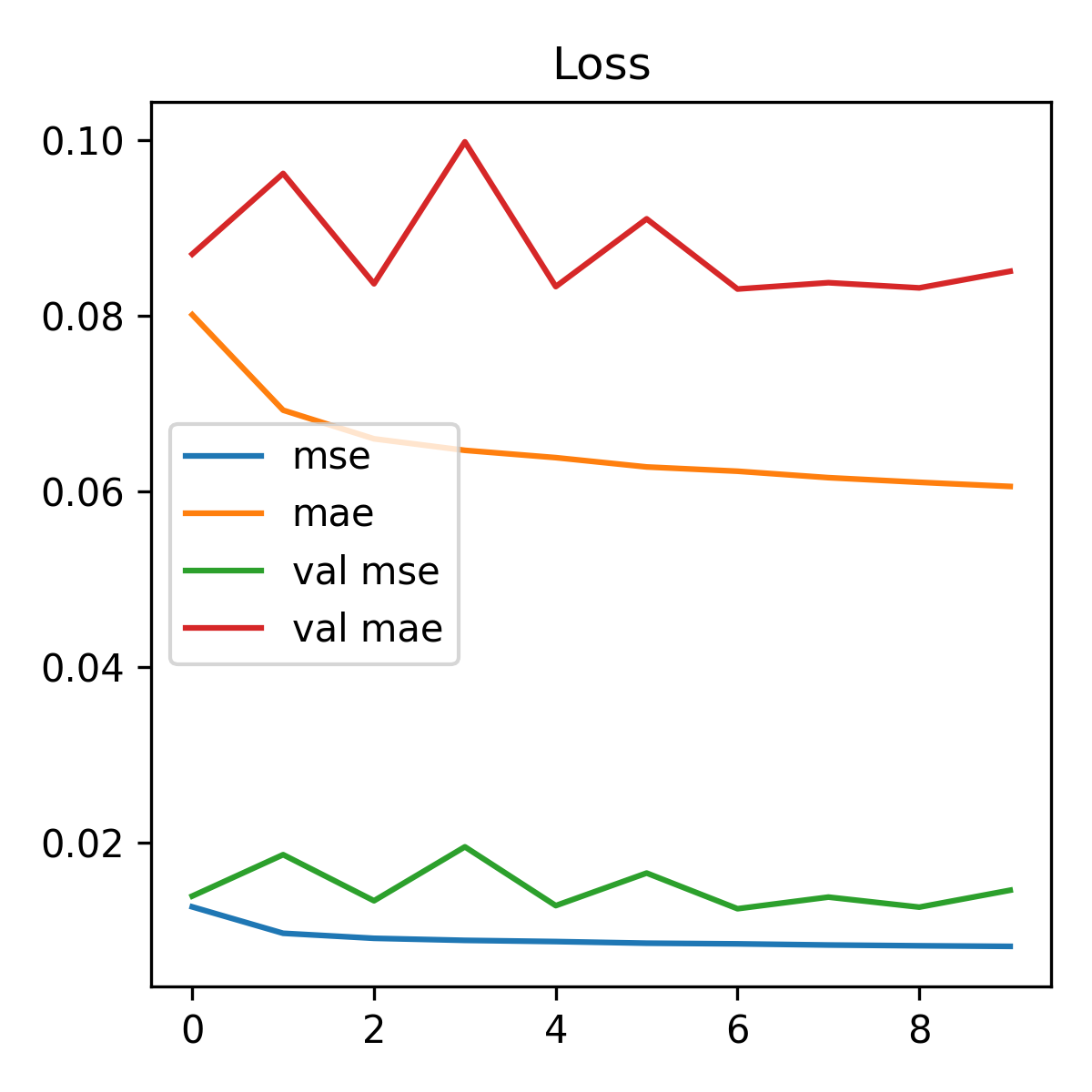}
	\caption{Performance metrics of the GRU model}
	\label{fig:gru}
\end{figure}

GRUs (Gated Recurring Unit) models are a modified version of the LSTM model. The enhancement in the GRU structure is the introduction of a forget gate. We have constructed a GRU model similar to that of the LSTM model, it has 3 layers - 2 GRU layers and 1 Dense layer. The GRU layers like the LSTM model, have 64 GRU cells each. The GRU model is also trained on the Adam optimizer with mean absolute error as the loss function. This model too has been trained for 10 epochs. The performance of the model has been tabulated in Table \ref{table:gru} and visualized in Figure \ref{fig:gru}.
\begin{table}[htbp]
\begin{center}
\centering
\begin{tabular}{|c|c|c|c|c|} 
\hline
    \textbf{Epochs} & \textbf{MSE} & \textbf{MAE} & \textbf{val. MSE} & \textbf{val. MAE}\\
    \hline
    1 & 0.0127 & 0.0801 & 0.0139 & 0.0870\\ 
    2 & 0.0097 & 0.0692 & 0.0186 & 0.0962\\ 
    3 & 0.0091 & 0.0660 & 0.0134 & 0.0836\\ 
    4 & 0.0089 & 0.0647 & 0.0195 & 0.0998\\ 
    5 & 0.0088 & 0.0638 & 0.0128 & 0.0833\\ 
    6 & 0.0086 & 0.0628 & 0.0166 & 0.0910\\
    7 & 0.0085 & 0.0623 & 0.0125 & 0.0830\\ 
    8 & 0.0083 & 0.0616 & 0.0138 & 0.0838\\ 
    9 & 0.0083 & 0.0610 & 0.0127 & 0.0832\\
    10 & 0.0082 & 0.0606 & 0.0146 & 0.0851\\
    \hline
\end{tabular}
\end{center}
\caption{The cost function(MSE - Mean Squared Error and MAE - Mean Absolute Error) for both training and validation for the GRU model.}
\label{table:gru}
\end{table}

\subsection{Optimized training of model}
The same LSTM model is used for optimized training using a Genetic Algorithm. 10 individuals are made by cloning the same model so that there is no favoritism in the training process. these 10 individuals are then trained for an epoch for a randomly assigned batch size. The loss obtained for each individual is then subtracted from the previous individual. For the first generation, we subtract it from zero. The individuals are sorted and passed over to the crossover phase. In the crossover phase, the models with the two highest losses are passed forward without any change. For the remaining 8 individuals, mating is performed between the individuals of the previous generation. These individuals go through the process of mutation based on a generated factor. Once the next generation is completely formed, it goes through the same process of training, crossover, and mutation. this process goes on for several generations. In our case, we have gone with 10 generations. Table \ref{table:optimized} shows the metrics of the last generation of individuals that we have obtained.

\begin{table}[htbp]
\begin{center}
\centering
\begin{tabular}{|c|c|c|c|c|} 
\hline
    \textbf{Individual} & \textbf{MSE} & \textbf{MAE} & \textbf{val. MSE} & \textbf{val. MAE}\\
    \hline
    1 & 0.0071 & 0.0582 & 0.0121 & 0.0805\\ 
    2 & 0.0085 & 0.0658 & 0.0130 & 0.0840\\ 
    3 & 0.0085 & 0.0658 & 0.0130 & 0.0840\\ 
    4 & 0.0082 & 0.0647 & 0.0128 & 0.0837\\ 
    5 & 0.0085 & 0.0672 & 0.0140 & 0.0865\\ 
    6 & 0.0085 & 0.0658 & 0.0130 & 0.0840\\
    7 & 0.0085 & 0.0658 & 0.0130 & 0.0840\\ 
    8 & 0.0085 & 0.0672 & 0.0152 & 0.0895\\ 
    9 & 0.0088 & 0.0643 & 0.0140 & 0.0850\\
    10 & 0.0088 & 0.0643 & 0.0140 & 0.0851\\
    \hline
\end{tabular}
\end{center}
\caption{The cost function(MSE - Mean Squared Error and MAE - Mean Absolute Error) for both training and validation for the GRU model.}
\label{table:optimized}
\end{table}

\subsection{Performance Analysis}
\begin{figure}[htbp]
	\centering
	\includegraphics[width=0.7\linewidth]{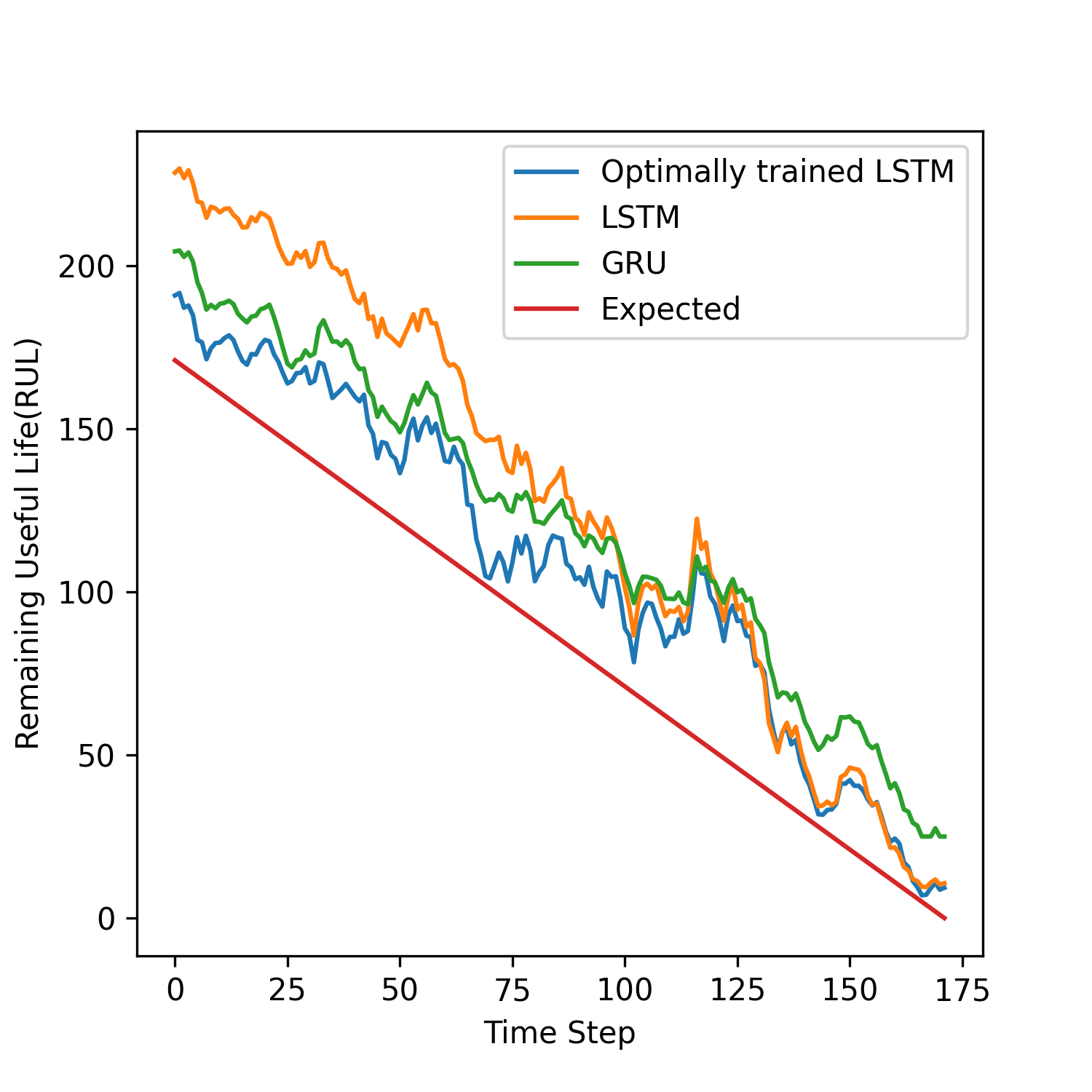}
	\caption{Predictions of the models against the actual remaining life of the jet engine}
	\label{fig:conv}
\end{figure}
Once the final generation of the optimized training model is obtained, the best individual from the lot is selected (this is done based on the least loss obtained by the model). This individual competes against the LSTM and GRU models that we have trained earlier. The LSTM and the GRU models show high variation and deviation from the expected values. The performance of the LSTM and GRU models are not too far off from the expected value but is overtaken by the genetically trained model. The genetically trained model correctly predicts the end of life and converges with the expected value in contrast to the LSTM and GRU model as shown in Figure \ref{fig:conv}. This model, therefore, serves as a better model for prediction and deployment.

\section{Conclusion and Future Work}
\label{Conclusion}
The main objective of Remaining Useful life Prediction is to obtain low-cost maintenance and conserve capital. The state-of-the-art implementations and models have shown great results in the domain of prognostics. Similarly, the proposed solution aims to optimize the training process of recurrent neural networks for better predictions. The results obtained can display the superiority of the optimized architecture over existing LSTM and GRU models. It can tune the existing models beyond manual limits with the help of cross-over and mutation. Current optimization is limited by the architecture and degree of randomization. The metric of $\Delta$ loss used can also be improvised upon. Further research would include greater tuning of hyper-parameters. Initially, all individuals represent the same architecture, but bring randomness to this could improve the efficiency of the algorithm in the later stages of training. Other deep learning architectures like CNNs and auto-encoders could be explored. Ensemble learning techniques could be used on the final generation to get average values.

\bibliographystyle{IEEEtran}
\bibliography{references}

\end{document}